\documentclass[letterpaper, 10 pt, conference]{ieeeconf}  

\IEEEoverridecommandlockouts                             
\usepackage{amsmath} 
\usepackage{amssymb}  
\usepackage{multirow}
\usepackage{graphicx}
\usepackage{tikz}
\usepackage{caption}
\usepackage{subcaption}
\usepackage{tikz,lipsum}
\usepackage[most]{tcolorbox}
\usepackage{graphicx}
\let\labelindent\relax
\usepackage{enumitem}

\title{\LARGE \bf
Self-Corrective Task Planning by Inverse Prompting \\
with Large Language Models
}

\author{Jiho Lee$^{1}$, Hayun Lee$^{1}$, Jonghyeon Kim$^{2}$, Kyungjae Lee$^{3}$, and Eunwoo Kim$^{1,*}$
\thanks{$^{1}$Jiho Lee, Hayun Lee, and Eunwoo Kim are with the School of Computer Science Engineering, Chung-Ang University, Seoul, 06974, Republic of Korea (e-mail: \{j2hoooo, hayun0406, eunwoo\}@cau.ac.kr).}
\thanks{$^{2}$Jonghyeon Kim is with the Department of AI, Chung-Ang University, Seoul, 06974, Republic of Korea (e-mail: kjh980610@cau.ac.kr).}
\thanks{$^{3}$Kyungjae Lee is with the Department of Statistics,  Korea University, Seoul, 02841, Republic of Korea (e-mail: kyungjae\_lee@korea.ac.kr).}
\thanks{* Corresponding author.}
}

\begin{document}

\maketitle
\thispagestyle{empty}
\pagestyle{empty}

%%%%%%%%%%%%%%%%%%%%%%%%%%%%%%%%%%%%%%%%%%%%%%%%%%%%%%%%%%%%%%%%%%%%%%%%%%%%%%%%
%%%%%%%%%%%%%%%%%%%%%%%%%%%%%%%%%%%%%%%%%%%%%%%%%%%%%%%%%%%%%%%%%%%%%%%%%%%%%%%%

\begin{abstract}
In robot task planning, large language models (LLMs) have shown significant promise in generating complex and long-horizon action sequences.
However, it is observed that LLMs often produce responses that sound plausible but are not accurate.
To address these problems, existing methods typically employ predefined error sets or external knowledge sources, requiring human efforts and computation resources.
Recently, self-correction approaches have emerged, where LLM generates and refines plans, identifying errors by itself.
Despite their effectiveness, they are more prone to failures in correction due to insufficient reasoning.
In this paper, we introduce InversePrompt, a novel self-corrective task planning approach that leverages inverse prompting to enhance interpretability. 
Our method incorporates reasoning steps to provide clear, interpretable feedback. 
It generates inverse actions corresponding to the initially generated actions and verifies whether these inverse actions can restore the system to its original state, explicitly validating the logical coherence of the generated plans.
The results on benchmark datasets show an average 16.3\% higher success rate over existing LLM-based task planning methods.
Our approach offers clearer justifications for feedback in real-world environments, resulting in more successful task completion than existing self-correction approaches across various scenarios.
\end{abstract}

%-----------------------------------------------------------------%
%
\begin{figure}[!tb]
    \centering
    \vspace{1.5mm}
    \begin{subfigure}[t]{0.475\linewidth}
        \centering    
        \includegraphics[width=\textwidth]{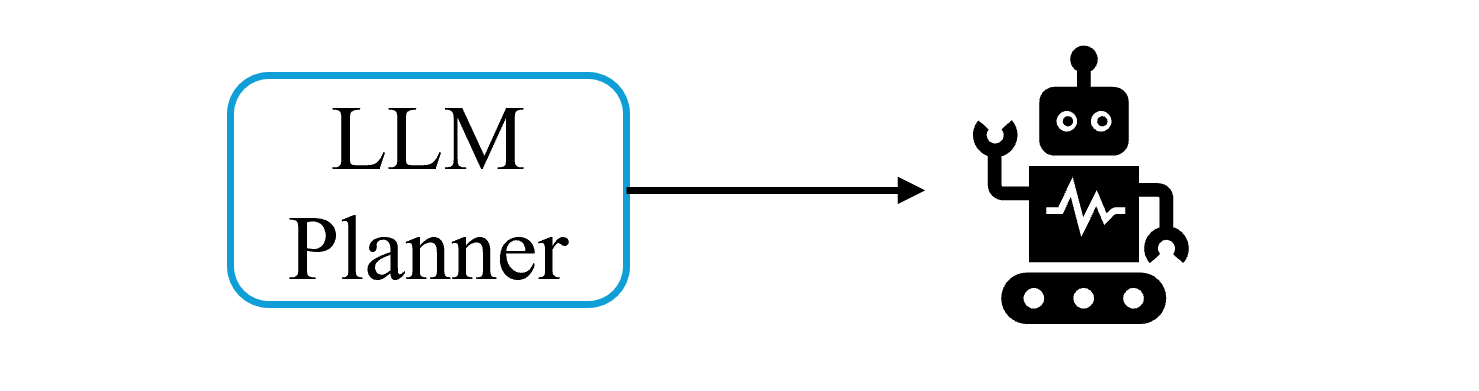}
        \caption{No validator}
        \label{subfig:fig_1_a}
    \end{subfigure}
    \hfill
    \begin{subfigure}[t]{0.475\linewidth}
        \centering    
        \includegraphics[width=\textwidth]{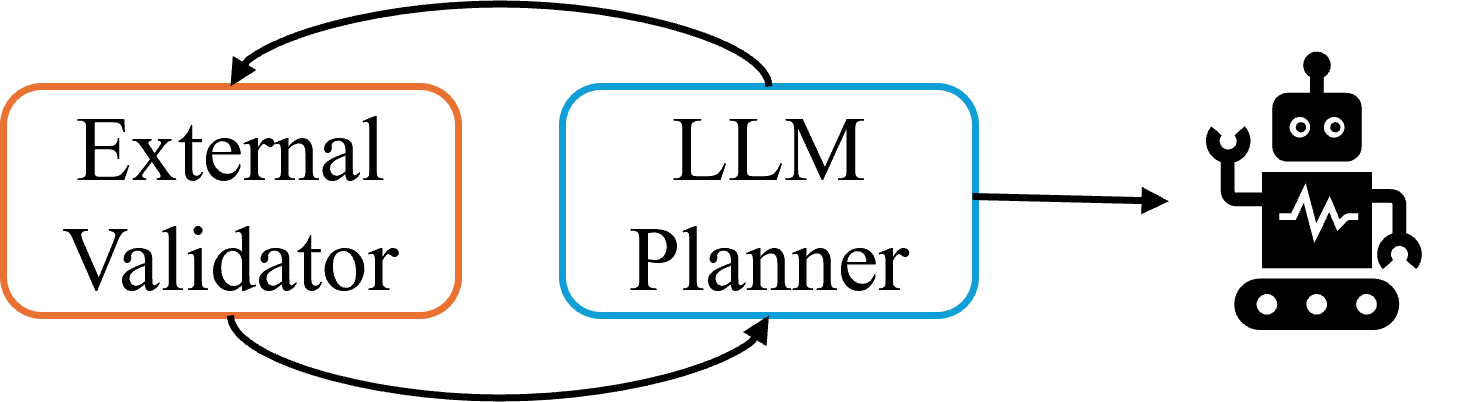}
        \caption{External validator}
        \label{subfig:fig_1_b}
    \end{subfigure}
    \\\vspace{2.6mm}
    \hfill
    \begin{subfigure}[b]{0.47\linewidth}
        \centering
        \includegraphics[width=\textwidth]{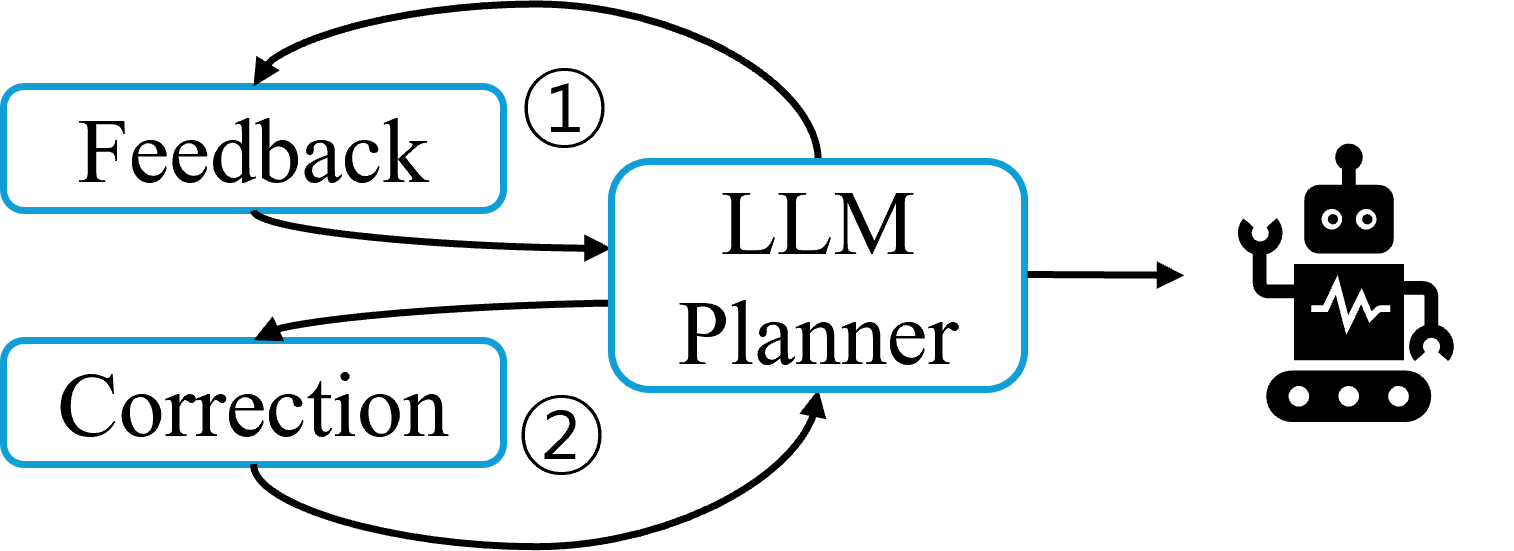}
        \caption{Self-correction\newline}
        \label{subfig:fig_1_c}
    \end{subfigure}
    \hfill
    \begin{subfigure}[b]{0.47\linewidth}
        \centering
        \includegraphics[width=\textwidth]{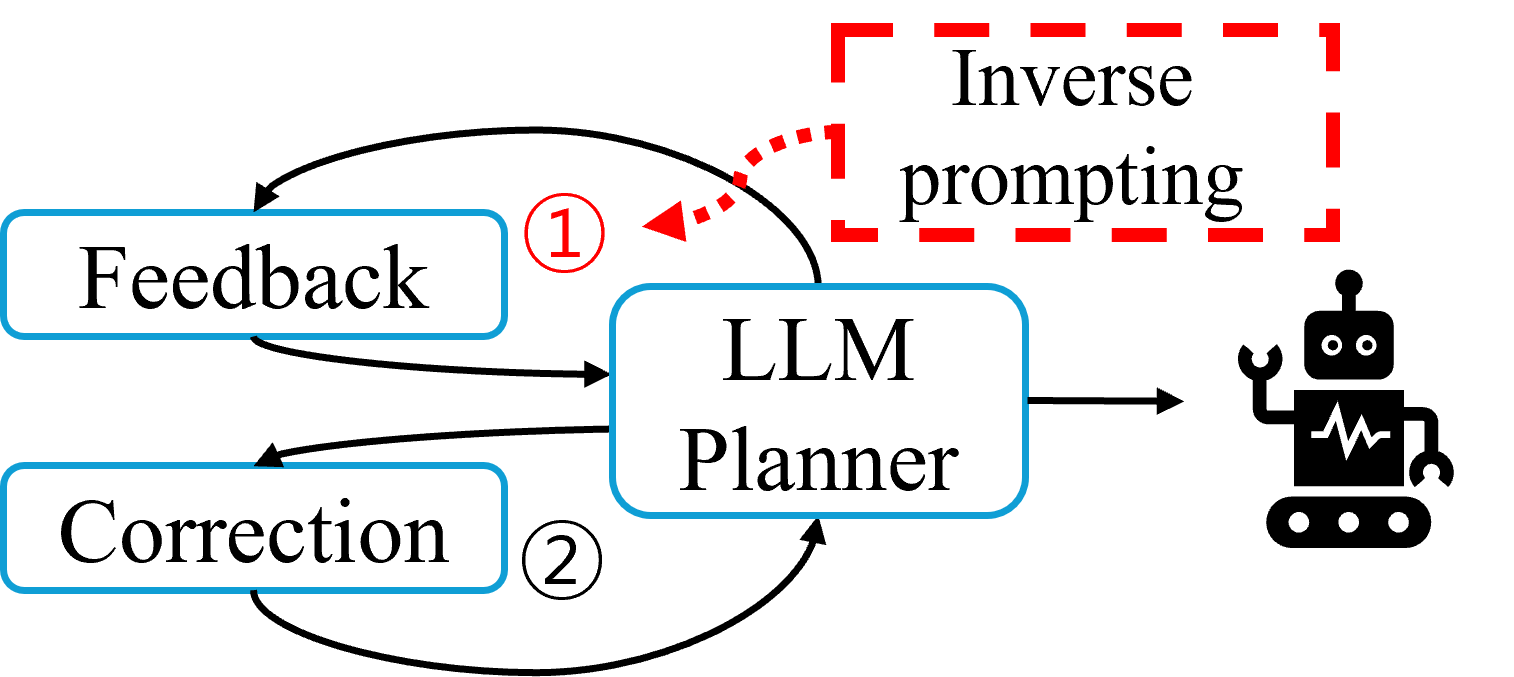}
        \caption{\centering{ Self-correction \newline with Inverse Prompting (Ours)}}
        \label{subfig:fig_1_d}
    \end{subfigure} 
    \\\vspace{1mm}
    \caption{ 
    Given a goal, the LLM planner generates an action sequence.
    Then, (a) a robot executes it without any validation.
    (b) The generated action sequence is validated by an external validator, which is constructed using rule-based methods or by retrieving knowledge from external sources.
    (c) The LLM planner validates and refines the generated action sequence through a self-correction process, providing feedback on its output and using it to correct actions.
    (d) The proposed method further enhances the self-correction process from (c) with inverse prompting.}
    \label{fig:fig_1}
\end{figure}
%
%-----------------------------------------------------------------%

%%%%%%%%%%%%%%%%%%%%%%%%%%%%%%%%%%%%%%%%%%%%%%%%%%%%%%%%%%%%%%%%%%%%%%%%%%%%%%%%
%%%%%%%%%%%%%%%%%%%%%%%%%%%%%%%%%%%%%%%%%%%%%%%%%%%%%%%%%%%%%%%%%%%%%%%%%%%%%%%%

\section{INTRODUCTION}
Recent advancements in large language models (LLMs) have demonstrated remarkable capabilities across various tasks, including translation~\cite{zhang2022opt}, code generation~\cite{nijkamp2023codegen}, arithmetic~\cite{lewkowycz2022solving}, and commonsense reasoning~\cite{brown2020language}.
Their ability to comprehend the context and generate complex outputs based on extensive knowledge about the world has led to growing attention in robot task planning, where precise decision-making and adaptability are crucial~\cite{brohan2023can,huang2023inner,liu2023llm+, shin2024task}.

However, LLMs continue to produce outputs that, while plausibly worded, are often inaccurate or infeasible~\cite{huang2023survey}.
This issue extends to robot task planning, where the models often generate task plans that may appear syntactically sound but are non-executable by robots~\cite{valmeekam2022large}.
To address these planning errors from LLMs, existing works have been investigated through two approaches: external validator and self-correction methods (see Fig.~\ref{fig:fig_1}(\subref{subfig:fig_1_b}) and \ref{fig:fig_1}(\subref{subfig:fig_1_c})).
The external validator method is widely employed with the predefined set of errors~\cite{skreta2023errors,joublin2024copal, raman2024cape} or external knowledge sources~\cite{liu2023reflect}.
~\cite{skreta2023errors,joublin2024copal, raman2024cape} construct a predefined set of error types and the corresponding feedback in advance to compare against the generated plan.
While effective for certain errors, it is constrained by their reliance on fixed rules, which may not address all possible errors or adapt to new scenarios.
It also requires substantial human efforts to construct strict rules.
To leverage external knowledge sources, we can retrieve knowledge from news articles, Wikipedia~\cite{guu2020retrieval,peng2023check}, or additional LLMs for verification~\cite{liu2023reflect}.
However, accessing these resources may not always be feasible due to high costs~\cite{barnett2024seven}.

To overcome the reliance on additional resources, a self-correction method~\cite{madaan2024self} has been applied to robot task planning~\cite{zhou2024isr, ming2023hicrisp}.
Their approaches enable LLM to identify and correct its mistakes, inspired by human problem-solving techniques~\cite{simon1962architecture}.
However, they simply adopt a self-correction method without considering the structured or symbolic characteristics of robot task planning.
As shown in Fig.~\ref{fig:fig_2}(b), they provide an explanation for the infeasibility of the action, but they lack thorough groundings or steps that support a clear reasoning process.
It is worth noting that existing single-step reasoning processes often fall short in long-horizon planning tasks, as demonstrated in~\cite{wei2022chain}.
It can potentially increase the likelihood of generating a less coherent response and degrade the capability of the feedback as a validator.
Consequently, the generated feedback is more prone to failures in correction, being both plausible-sounding and inaccurate.

In this paper, we introduce a self-corrective planning approach with inverse prompting, called InversePrompt, to incorporate multiple reasoning steps, inspired by the inversion principle in mathematical verification~\cite{canobi2005children}.
The inverse prompting is designed to generate the conclusion with three reasoning steps, as highlighted by the black underline in Fig.~\ref{fig:fig_2}(c).
It firstly produces inverse actions and the corresponding states.
Note that the inverse action can be generated based on the symbolic and linguistic characteristics of the action language.
It then verifies whether the inverted state can return to the original state.
The result of an operation, such as $x + y = z$, is validated by reversing it, for example, $z - y = x$.
In this context, the inverse action and the original state correspond to the role of $-y$ and $x$, respectively.
The proposed method easily generates more detailed and accurate feedback by analyzing the difference between the inverted and the original states.
In this way, the proposed InversePrompt effectively enhances the reliability of the feedback while validating the logical flow of the generated plans.

We validate the proposed method under extensive benchmark scenarios, including Ballmoving~\cite{ silver2024generalized}, Blocksworld~\cite{liu2023llm+}, and Cooking~\cite{zhou2024isr} environments.
The results show that the proposed method outperforms existing standard prompting and external validator methods by a margin of 17.5\% and 4.2\% on average, respectively.
We also evaluate the proposed method in real-world environments with infeasible plans.
The proposed approach demonstrates its effectiveness in error correction, resulting in more successful task completion than existing self-correction methods.
The contributions of this work are as follows:
%-----------------------------------------------------------------%
%
\begin{itemize}
    \item We propose self-corrective planning of tasks with an inverse prompting strategy, named InversePrompt, enabling an LLM to provide explicit justifications in the reasoning process.
    \item The proposed approach introduces a multi-step reasoning process for generating comprehensive feedback and explicitly validating the logical flow of the generated plans with inverse actions and the states.
    \item  Extensive experimental results demonstrate that the proposed approach achieves a 16.3\% higher average success rate with fewer attempts than other competitors across benchmark datasets and effectively handles unexpected failures in real experiments.
\end{itemize}
%
%-----------------------------------------------------------------%

%-----------------------------------------------------------------%
%
\begin{figure}[t!]
    \centering
    {\small 
    \begin{tcolorbox}[colback=gray!0, colframe=black!60, title=(a) Question, left=1mm, right=1mm, top=0mm, bottom=0mm]
    \begin{itemize}[leftmargin=*]
        \item \textbf{Current  \hspace{-1mm}state:} \hspace{-1mm}{\texttt{\{(robot-at \hspace{-1mm}robot1 \hspace{-1mm}room1)\hspace{-1mm}(at \hspace{-1mm}ball1 \hspace{-1mm}room2)(at ball2 room3)(at ball3 room1)(at ball4 room2)\}}}
        \item \textbf{Goal \hspace{-1mm}state:} \hspace{-1mm}{\texttt{\{(at\hspace{-1mm} ball1\hspace{-1mm} room1)\hspace{-1mm}(at\hspace{-1mm} ball2\hspace{-1mm} room1)\hspace{-1mm}(at ball3 room3)(at ball4 room4)\}}}
        \item \textbf{Examined action:} {\texttt{(pick ball1 room2)}}
    \end{itemize}
    \end{tcolorbox}}
    {\small    
    {\begin{tcolorbox}[colback=gray!0, colframe=black!60, left=1mm, right=1mm, top=0mm, bottom=0mm, title=(b) Answer with standard prompting]
    \begin{itemize}[leftmargin=*]
        \item \textbf{Resulting state:} {\texttt{\{(arm-ball1)(robot-at robot1 room2)(at ball1 room2)(at ball2 room3)(at ball3 room1)(at ball4 room2)\}}}
        \item \textbf{The action is} {{wrong {\color{red}\underline{because ball1 is not in room2.}}}}
    \end{itemize}
    \end{tcolorbox}}}
    {\small
    \begin{tcolorbox}[colback=gray!0, colframe=black!60, left=1mm, right=1mm, top=0mm, bottom=0mm,  title= (c) Answer with inverse prompting (Ours)]
    \begin{itemize}[leftmargin=*]
        \item\textbf{Resulting  state:} {\texttt{\{(arm-ball1)(robot-at robot1 room2)(at ball1 room2)(at ball2 room3)(at ball3 room1)(at ball4 room2)\}}}
        \item\textbf{\underline{Inverse action}:} {\texttt{(drop ball1 room2)}}
        \item\textbf{\underline{Inversed state}:} {\texttt{\{(robot-at robot1 room2)(at ball1 room2)(at ball2 room3)(at ball3 room1)(at ball4 room2)\}}}
        \item\textbf{\underline{Difference}:} \hspace{-3mm}{\texttt{(robot-at\hspace{-1mm} {\color{red}\underline{robot1\hspace{-1mm} room2}}) \hspace{-1mm}\&\hspace{-1mm} (robot-at \hspace{-1mm}{\color{red}\underline{robot1\hspace{-1mm} room1}})}}
        \item\textbf{The action is} {{wrong {\color{red}\underline{because robot1 is in room1, not room2.}}}}
    \end{itemize}
    \end{tcolorbox}}
    \caption{ 
    Examples of the generated feedback with the standard prompting and the proposed inverse prompting under the Ballmoving domain, given the question in (a).
    The text underlined in red indicates the groundings for the final answer, while the text underlined in black highlights the reasoning steps in our approach.}
    \label{fig:fig_2}
\end{figure}
%
%-----------------------------------------------------------------%

%%%%%%%%%%%%%%%%%%%%%%%%%%%%%%%%%%%%%%%%%%%%%%%%%%%%%%%%%%%%%%%%%%%%%%%%%%%%%%%%
%%%%%%%%%%%%%%%%%%%%%%%%%%%%%%%%%%%%%%%%%%%%%%%%%%%%%%%%%%%%%%%%%%%%%%%%%%%%%%%%

\section{RELATED WORKS}
%%%%%%%%%%%%%%%%%%%%%%%%%%%%%%%%%%%%%%%%%%%%%%%%%%%%%%%%%%%%%%%%%%%%%%%%%%%%%%%%
\subsection{Task Planning by LLMs}
Large language models (LLMs), due to their capacity to handle complex sequences with rich commonsense knowledge, have been increasingly utilized to generate long-horizon action sequences for robot task planning~\cite{brohan2023can, huang2023inner,liu2023llm+, shin2024task, firoozi2023foundation}.
\cite{brohan2023can} produces the action sequences by grounding the given task and current environment within the predefined set of skills.
\cite{huang2023inner} leverages the scene description from pre-trained models or humans while generating action sequences.
\cite{liu2023llm+} proposes the translator and planner with LLMs to produce action sequences by converting natural language instructions into the planning domain definition language (PDDL) formulation~\cite{haslum2019introduction}.
\cite{shin2024task} presents an LLM-based planner that generates action sequences while automatically producing trainable datasets from language instructions without human supervision.
These approaches do not account for potential failures during the planning process, focusing solely on generating action sequences with LLMs, even if LLMs often produce plausible-sounding but inaccurate plans~\cite{valmeekam2022large}.
To address these issues, the proposed method is designed to perform self-corrective task planning through an enhanced prompt strategy. 
This identifies and corrects errors by itself while providing more comprehensive feedback.

%%%%%%%%%%%%%%%%%%%%%%%%%%%%%%%%%%%%%%%%%%%%%%%%%%%%%%%%%%%%%%%%%%%%%%%%%%%%%%%%
\subsection{Corrective Planning}
Despite the effectiveness of LLMs in generating task plans, their limitations become evident when the generated plans are occasionally not feasible in practice~\cite{valmeekam2022large}.
To refine the original task plans, existing works mostly adopt external validators such as pre-defined rules~\cite{skreta2023errors, joublin2024copal, raman2024cape} or additional LLMs~\cite{liu2023reflect}.
\cite{skreta2023errors} proposes an iterative prompting strategy employing a rule-based verifier as a syntax checker and a static analyzer.
\cite{joublin2024copal} introduces a multi-level feedback loop with pre-defined error types.
\cite{raman2024cape} addresses failures in execution caused by precondition violations through the use of corrective prompts.
\cite{liu2023reflect} proposes a multi-modal hierarchical summary of the execution history to detect and correct failures with additional LLMs.

Recently, \cite{zhou2024isr,ming2023hicrisp} have adopted the self-correction method in robot task planning, where the planner figures out and refines errors by itself. 
These works using LLMs for error detection and correction rely on a na\"ive reasoning process through standard prompting, directly answering feasibility or potential problems with a single reasoning step.
The absence of thorough reasoning or explicit justifications for responses can lead to failures in detecting errors or the generation of unhelpful feedback for correction. 
In this paper, we introduce an inverse prompting strategy for the self-correction process that involves a multi-step reasoning approach over generated task plans and explicit groundings through inverse actions.

%%%%%%%%%%%%%%%%%%%%%%%%%%%%%%%%%%%%%%%%%%%%%%%%%%%%%%%%%%%%%%%%%%%%%%%%%%%%%%%%
%%%%%%%%%%%%%%%%%%%%%%%%%%%%%%%%%%%%%%%%%%%%%%%%%%%%%%%%%%%%%%%%%%%%%%%%%%%%%%%%

\section{METHOD}
%%%%%%%%%%%%%%%%%%%%%%%%%%%%%%%%%%%%%%%%%%%%%%%%%%%%%%%%%%%%%%%%%%%%%%%%%%%%%%%%
\subsection{Framework}
\begin{figure}[t!]
    \centering
    \includegraphics[width=\columnwidth]{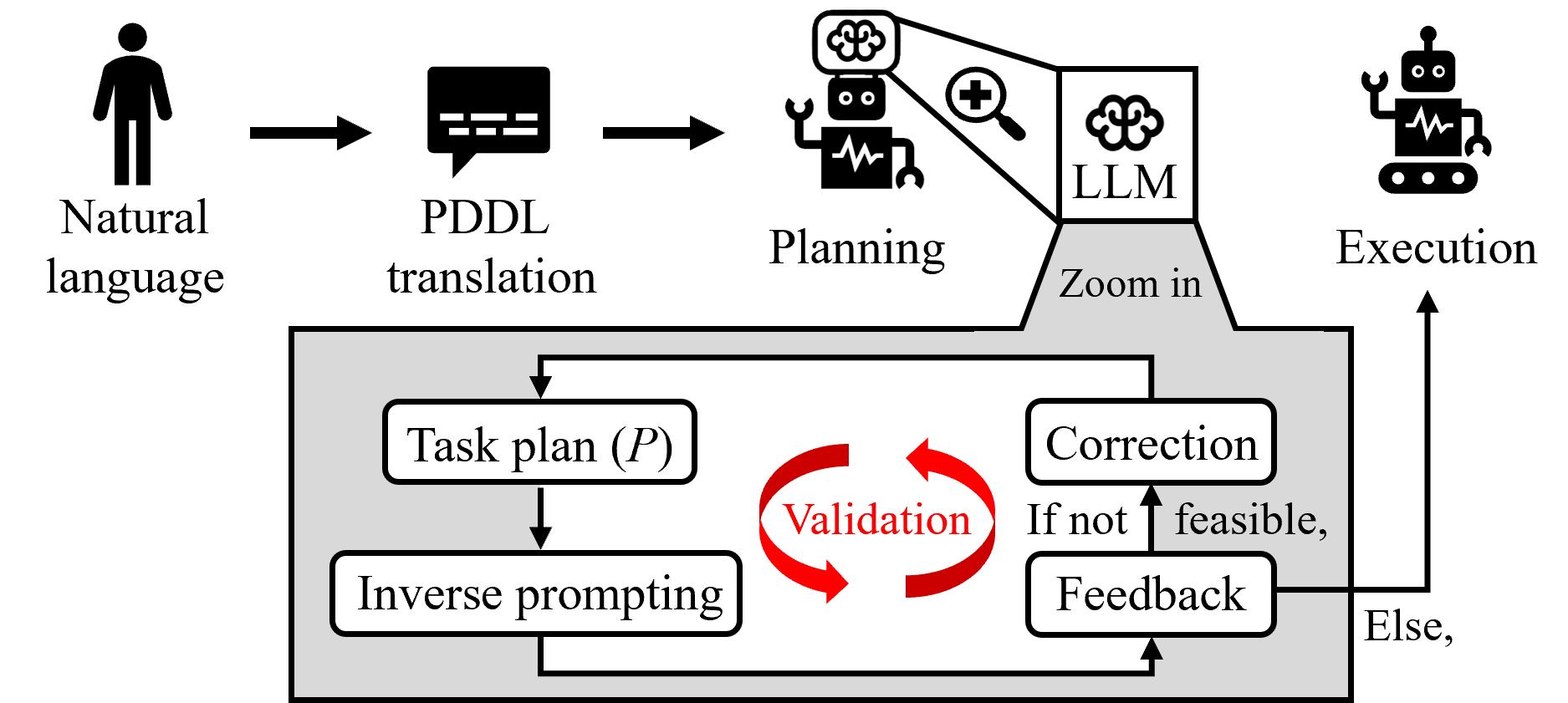}
    \caption{     
    An overview of the proposed overall process.}
    \label{fig:fig_3}
\end{figure}

In this work, we propose a method for enabling an LLM to self-correct errors it produces, including both self-detection and self-rectification, without the need for extra resources~\cite{skreta2023errors,joublin2024copal,raman2024cape,liu2023reflect}.
As illustrated in Fig.~\ref{fig:fig_3}, we first translate the given task goal described in natural language into a PDDL formulation~\cite{haslum2019introduction} using an LLM, where actions are predefined with their corresponding robot skills~\cite{liu2023llm+, zhou2024isr}.
This formulation consists of the initial state $s_{0}$ and the goal state $s_{goal}$, representing the relevant environmental information.
Then, the LLM generates a sequence of task plans $P=\{(s_{1}, a_1), ..., ( s_{T}, a_T)\}$, where the goal state can be achieved as $s_{goal}=s_{T}$ starting from $s_{0}$.
This sequence comprises sets of actions $\mathcal{A}=\{a_1, ..., a_T\}$ and states $\mathcal{S}=\{s_{1}, ..., s_{T}\}$.
The state transitions can be represented as
%-----------------------------------------------------------------%
%
\begin{equation}
    s_{0} + a_1 = s_1, \;\; \\
    \cdots, \;\;   \\
    s_{T-1} + a_T = s_{T},
    \label{eq:eq_1}
\end{equation}
%
%-----------------------------------------------------------------%
where the addition operation symbolically represents the transformation from one state to the next through an action.

The generated plan $P$ undergoes a self-correction process, in which the LLM evaluates the feasibility and correctness of the actions and states itself. 
If the plan is found to be infeasible, feedback is generated to rectify errors in the re-planning phase. 
This self-correction process with re-planning is repeated until either a feasible action sequence is identified or a predefined number of repetitions is reached. 
We introduce an inverse prompting strategy, called InversePrompt, to enable the LLM to produce explicit feedback with multi-step reasoning during the self-correction process.
It is designed to generate inverse actions and evaluate whether the current environment can be restored through the inverse actions, providing justifications for the responses and explicitly validating the logical flow of plans.

Note that we iteratively generate and validate each step of the plan based on the current state information, allowing us to address unexpected failures that often occur in real-world scenarios.
The actions and states generated from the self-correction process can be compared with the actual state observed after the execution of the robot.
This comparison determines whether the original plan needs to be refined and guides the subsequent execution.

%-----------------------------------------------------------------%
%
\begin{table*}[t!]
    \caption{
    Comparison results of success rates across various scenarios and models.
    Self-corr. w/o IP denotes the self-correction method without the proposed inverse prompting (IP).
    The bold text indicates the best results.}
    \centering
    {\resizebox{\textwidth}{!}{%
    \begin{tabular}{cc|cccc|cccc}
        \hline
        \multicolumn{2}{c|}{\multirow{3}{*}{Scenario}} & \multicolumn{4}{c|}{GPT-4o-mini} & \multicolumn{4}{c}{Gemini-1.5-Flash} \\
        \cline{3-10} 
        \multicolumn{2}{c|}{}& \begin{tabular}[c]{@{}c@{}}No\\ Validator\end{tabular} & \begin{tabular}[c]{@{}c@{}}External\\ Validator\end{tabular} & \begin{tabular}[c]{@{}c@{}}Self-corr.\\ w/o IP\end{tabular} & \begin{tabular}[c]{@{}c@{}}InversePrompt\\(Ours)\end{tabular} & \begin{tabular}[c]{@{}c@{}}No\\ Validator\end{tabular} & \begin{tabular}[c]{@{}c@{}}External\\ Validator\end{tabular} & \begin{tabular}[c]{@{}c@{}}Self-corr.\\ w/o IP\end{tabular} & \begin{tabular}[c]{@{}c@{}}InversePrompt\\(Ours)\end{tabular} \\
        \hline
        \multirow{2}{*}{Ballmoving} & $N$=3 & 70\% & 85\% & 85\% & \textbf{100\%} & 20\% & 90\% & 70\% & \textbf{100\%} \\
        & $N$=4 & 50\% & 75\% & 75\% & \textbf{100\%} & 25\%  & \textbf{55\%} & 25\% &  50\% \\
        \hline
        \multirow{2}{*}{Blocksworld} & $N$=3 & 75\% & \textbf{90\%} & 75\% & \textbf{90\%} & 70\% & \textbf{95\%}  & 85\% & \textbf{95\%} \\
        & $N$=4 & 55\% & 60\% & 50\% & \textbf{65\%} & 85\% & \textbf{ 100\%} & 85\%  & \textbf{100\%} \\
        \hline
        \multirow{2}{*}{Cooking} & $N$=3 & 45\% & {65\%}  & 50\%  & \textbf{70\%}  &  80\% & \textbf{95\%} & 80\% &   {90\%} \\
        & $N$=4  & 30\% &  \textbf{45\%} &  30\% & \textbf{45\%}& 75\%  & \textbf{100\%} & 70\% &  \textbf{100\%} \\
        \hline
        \multicolumn{2}{c|}{Average} &  54\%  & 70\%  & {61\%}  &  \textbf{78\%} &  {59\%} & \textbf{89\%} &  {72\%} &\textbf{89\%} \\                
        \hline
    \label{tab:table_1}
    \end{tabular}}}
\end{table*}
%
%-----------------------------------------------------------------%

%%%%%%%%%%%%%%%%%%%%%%%%%%%%%%%%%%%%%%%%%%%%%%%%%%%%%%%%%%%%%%%%%%%%%%%%%%%%%%%%
\subsection{Inverse Prompting}
To facilitate the generation of justifications within its responses during self-correction by an LLM, we introduce an inverse prompting strategy.
This approach aims to enhance the capability of an LLM to engage in multi-step reasoning, in contrast to the previous methods that typically perform the reasoning process in a single step.
It involves three primary steps: 1) verifying the generated state, 2) applying inverse actions, and 3) confirming the goal state.
Note that a key difference of this approach compared to existing methods~\cite{zhou2024isr,ming2023hicrisp} lies in the reasoning by the second step, validation with inverse actions.

In the first step, state verification, the state $s_t$ for each action $a_t$ at time step $t$ in the generated sequence is validated during the planning phase. 
This step ensures that the generated action is appropriate for achieving the desired state and verifies whether the resulting state is accurate based on the executed action.
Second, the inverse action is generated using inverse prompting and applied to the validated state from the first step, resulting in an inverted state.
The feasibility of the action is evaluated by comparing the inverted state with the original state.
Last, the goal state verification confirms whether the final state $s_T$ matches the pre-determined goal state $s_{goal}$. 
%-----------------------------------------------------------------%
%
%
%-----------------------------------------------------------------%

In task planning, particularly described in the PDDL formulation, actions such as ``pick" and ``putdown" are inverses of each other, and so are ``stack" and ``unstack."
The states that result from these actions have specific consequences.
For example, in the Blocksworld domain~\cite{liu2023llm+}, which involves stacking blocks into a specified order, after the ``pick" action, the object should no longer be on the table, while after a ``putdown" action, it is placed back on the table.
Considering these aspects, our proposed method assesses the feasibility of a given plan by verifying whether the environment can return to its original state after performing the inverse action.

The inverse prompting involves three steps to produce the final response with clear justifications.
It firstly generates the inverse of a given action, $(a_t)^{-1}$, and then applies it to the current state to verify the possibility of returning to the original state.
If executing the inverse action returns to the original state as $s_{t} + (a_t)^{-1} = s_{t-1}$, the action is considered feasible.
Conversely, if it fails to return as $s_{t} + (a_t)^{-1} \neq s_{t-1}$, it provides feedback to correct the errors.
To this end, the LLM compares the original state $s_{t-1}$ with the inverted state after applying the inverse action $\hat{s}_{t-1}=s_{t} + (a_t)^{-1}$.
The difference is used as feedback to guide the re-planning phase.
This feedback generated from comparing the two states allows for more explicit corrections for re-planning.
Additionally, the reasoning steps allow the LLM to produce richer feedback while progressively generating its justifications and validating the logical flow of plans during self-correction.

Note that during the execution phase, the progress of execution can be monitored through self-correction by comparing the states in the plan $\mathcal{S}=\{s_{1}, ..., s_{T}\}$ with the observed states.
If discrepancies arise between the planned and observed states, these differences are utilized as feedback to the planner.
This process allows the planner to revise infeasible plans, guiding the development of a more feasible and logically coherent plan.

%%%%%%%%%%%%%%%%%%%%%%%%%%%%%%%%%%%%%%%%%%%%%%%%%%%%%%%%%%%%%%%%%%%%%%%%%%%%%%%%
%%%%%%%%%%%%%%%%%%%%%%%%%%%%%%%%%%%%%%%%%%%%%%%%%%%%%%%%%%%%%%%%%%%%%%%%%%%%%%%%

\section{Experiments}
We validated the effectiveness of the proposed method, InversePrompt, with an extensive set of experiments conducted in benchmark environments (Section~\ref{sec:benchmark}) and real-world settings (Section~\ref{sec:real_world_exp}).
The experiments were designed to assess two objectives: 1) the impact of the proposed inverse prompting in self-correction across scenarios and models and 2) the capability to handle unexpected situations during execution by detecting and refining the plan accordingly.

%%%%%%%%%%%%%%%%%%%%%%%%%%%%%%%%%%%%%%%%%%%%%%%%%%%%%%%%%%%%%%%%%%%%%%%%%%%%%%%%
\subsection{Benchmark Experiments}
\label{sec:benchmark}
%%%%%%%%%%%%%%%%%%%%%%%%%%%%%%%%%%%%%%%%%%%%%%%%%%%%%%%%%%%%%%%%%%%%%%%%%%%%%%%%
\subsubsection{Setup}
We conducted experiments using three widely adopted benchmarks, Ballmoving~\cite{silver2024generalized}, Blocksworld~\cite{liu2023llm+}, and Cooking~\cite{zhou2024isr}.
In the Ballmoving scenario, the objective is to redistribute $N$ balls, initially placed randomly across four rooms, into their goal locations with the constraint that only one ball can be moved at a time.
The Blocksworld scenario requires stacking $N$ blocks, initially placed randomly on the table, into a specified order by moving a single block at a time.
The Cooking scenario involves distributing up to six different ingredients into $N$ pots according to the recipes.
If an ingredient is picked, it can then be distributed into several pots.
We generated 20 randomized test cases for each environment, each with a distinct initial state and a goal state.
Following~\cite{zhou2024isr}, we evaluated the performance on scenarios of varying complexity by setting $N$ to 3 and 4, respectively. 
The mean lengths of task sequences for the scenarios with $N=3$ and $N=4$ are 10 and 13 for Ballmoving, 5 and 8 for Blocksworld, and 20 and 23 for Cooking, respectively.

We utilized two models to verify the applicability of the proposed method: GPT-4o-mini~\cite{gpt-4o-mini}, with a model size of approximately 8 billion parameters, and Gemini-1.5-Flash~\cite{reid2024gemini}, which has a larger model size than GPT-4o-mini although details are not publicly disclosed.
For GPT-4o-mini, we set the temperature, which controls the randomness of the response, to 0 for translation and incremented it by 0.1 up to 0.4 with each refinement step in the self-correction process, including planning and validation.
For Gemini-1.5-Flash, we set the temperature to 1 for translation and increased it incrementally by 0.1 up to 1.4 with each refinement step during planning, and 2 for validation.
We applied few-shot in-context learning~\cite{brown2020language}, a widely used technique, by providing prompt examples to guide the response of LLM.
The success or failure of the generated plan was assessed using a rule-based simulation system used in~\cite{zhou2024isr}.
The maximum number of refinement steps was set to 10.

%%%%%%%%%%%%%%%%%%%%%%%%%%%%%%%%%%%%%%%%%%%%%%%%%%%%%%%%%%%%%%%%%%%%%%%%%%%%%%%%
\subsubsection{Comparison methods}
We conducted comparative experiments with three types of approaches.
Firstly, we compared the proposed method to the baseline approach that does not employ any validator~\cite{brohan2023can, huang2023inner}, named No validator, performing a single planning stage with an LLM.
Secondly, we employed a rule-based external validator that evaluates pre-defined conditions for each action~\cite{skreta2023errors,joublin2024copal,raman2024cape}.
It detects errors and generates feedback based on pre-defined rules.
Lastly, we compared with the self-correction approach using a standard prompting strategy~\cite{zhou2024isr,ming2023hicrisp}.

%%%%%%%%%%%%%%%%%%%%%%%%%%%%%%%%%%%%%%%%%%%%%%%%%%%%%%%%%%%%%%%%%%%%%%%%%%%%%%%%
\subsubsection{Results}
As in TABLE~\ref{tab:table_1}, we summarize the comparison results on the Ballmoving, Blocksworld, and Cooking scenarios using the two models.
We can observe that InversePrompt significantly outperforms other methods, achieving average success rates of 78\% and 89\% with GPT-4o-mini and Gemini-1.5-Flash, respectively. 
This represents an improvement of 24\% and 30\% over the baseline that does not employ a validator.
Compared to the external validator, the proposed approach achieves up to 8\% higher average success rates across models and scenarios.
It indicates that its rules for error detection and feedback, although strictly constructed by human experts, may be less adaptable or interpretable.
Additionally, the method employing standard prompting in self-correction exhibits, on average, a 17\% lower accuracy compared to our InversePrompt for all scenarios and models.
The proposed inverse prompting method enables the planner to complete more tasks through the presented self-correction.

%-----------------------------------------------------------------%
%
\begin{table}[t!]
    \small
    \caption{Performance of the self-correction process.}
    \begin{tabular}{cc|cc|cc}
        \hline
        \multicolumn{2}{c|}{\multirow{3}{*}{Scenario}} & \multicolumn{2}{c|}{GPT-4o-mini} & \multicolumn{2}{c}{Gemini-1.5-Flash} \\
        \cline{3-6} 
        \multicolumn{2}{c|}{}& \begin{tabular}[c]{@{}c@{}}Self-corr.\\ w/o IP\end{tabular} & \begin{tabular}[c]{@{}c@{}} Ours\end{tabular} & \begin{tabular}[c]{@{}c@{}}Self-corr.\\ w/o IP\end{tabular} & \begin{tabular}[c]{@{}c@{}}Ours\end{tabular} \\ 
        \hline
        \multirow{2}{*}{Ballmoving} & $N$=3 & 80\% & \textbf{100\%} & 70\% & \textbf{85\%} \\
        & $N$=4 & 80\% & \textbf{100\%} & 25\% & \textbf{55\%} \\
        \hline
        \multirow{2}{*}{Blocksworld} & $N$=3 & 30\% & \textbf{35\%}  & 90\% & \textbf{100\%} \\
        & $N$=4 & \textbf{60\%} & \textbf{60\%} & 85\%  & \textbf{100\%}  \\
        \hline
        \multirow{2}{*}{Cooking} & $N$=3  & 55\%  & \textbf{70\%} & 80\% & \textbf{85\%} \\
         & $N$=4 & 30\% & \textbf{35\%} & 85\%  & \textbf{100\%}  \\ 
         \hline
        \multicolumn{2}{c|}{Average} & 56\% & \textbf{67\%} & 73\% & \textbf{88\%} \\
        \hline
    \label{tab:table_2}
    \end{tabular}
\end{table}
%
%-----------------------------------------------------------------%
We assessed the performance of the self-correction process to determine whether it accurately identified successes and failures in TABLE~\ref{tab:table_2}.
The accuracy was computed as $(TP+TN)/M\times100$, where $TP$ and $TN$ denote the number of true positives (identifying an actual successful plan as success) and true negatives (classifying a failed plan as failure), respectively, and $M$ is the number of test cases.
Our method applying inverse prompting demonstrates approximately 11\% and 16\% higher accuracy compared to the standard prompting approach without the proposed inverse prompting, for the GPT-4o-mini and Gemini-1.5-Flash, respectively, across all scenarios.
Notably, ours exhibits an impressive performance improvement of 30\% in the Ballmoving scenario with four balls on Gemini-1.5-Flash compared to the standard reasoning process.
This improvement indicates that the proposed inverse prompting method enhances the ability of LLM, improving the self-correction process.
In contrast, the method without inverse prompting often fails to detect errors, mistakenly classifying them and resulting in lower performance.

\begin{figure}[t!]
    \centering
    \begin{subfigure}[b]{1.0\linewidth}
        \centering
        \includegraphics[width=\columnwidth]{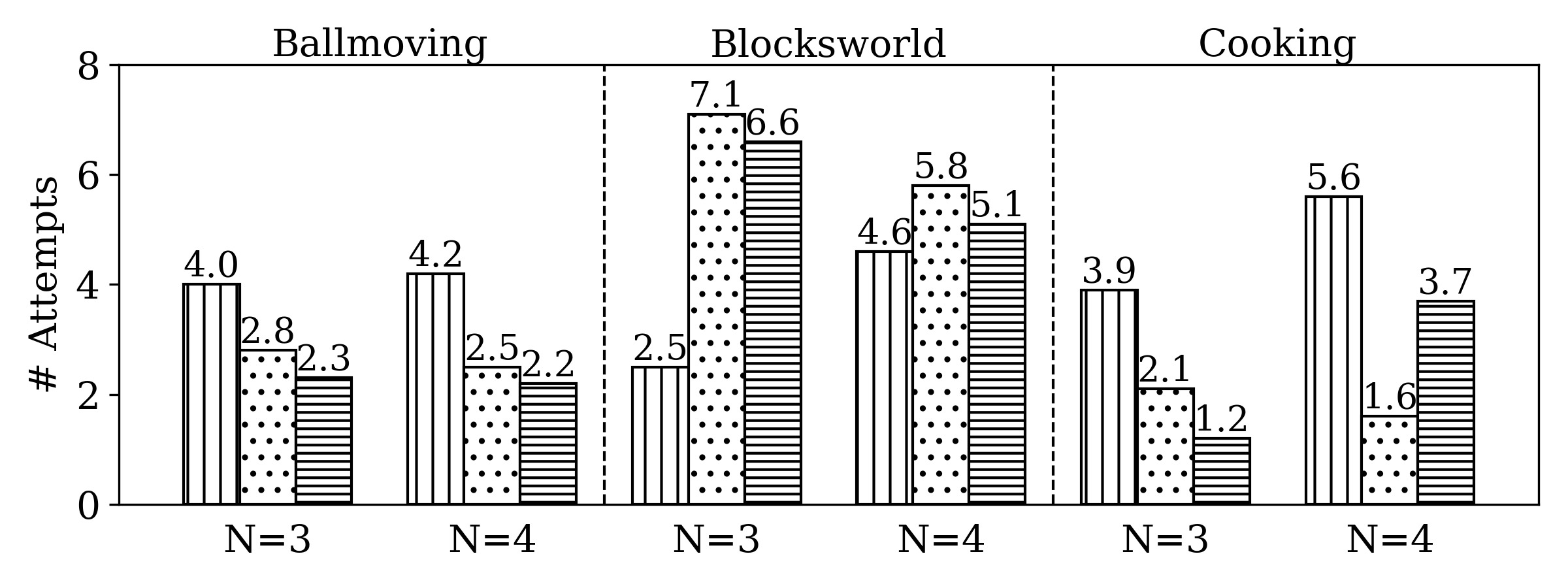}
        \caption{GPT-4o-mini}
        \label{subfig:fig_4_a}
    \end{subfigure}
    \\\vspace{2.5mm}
    \begin{subfigure}[b]{1.0\linewidth}
        \centering
        \includegraphics[width=\columnwidth]{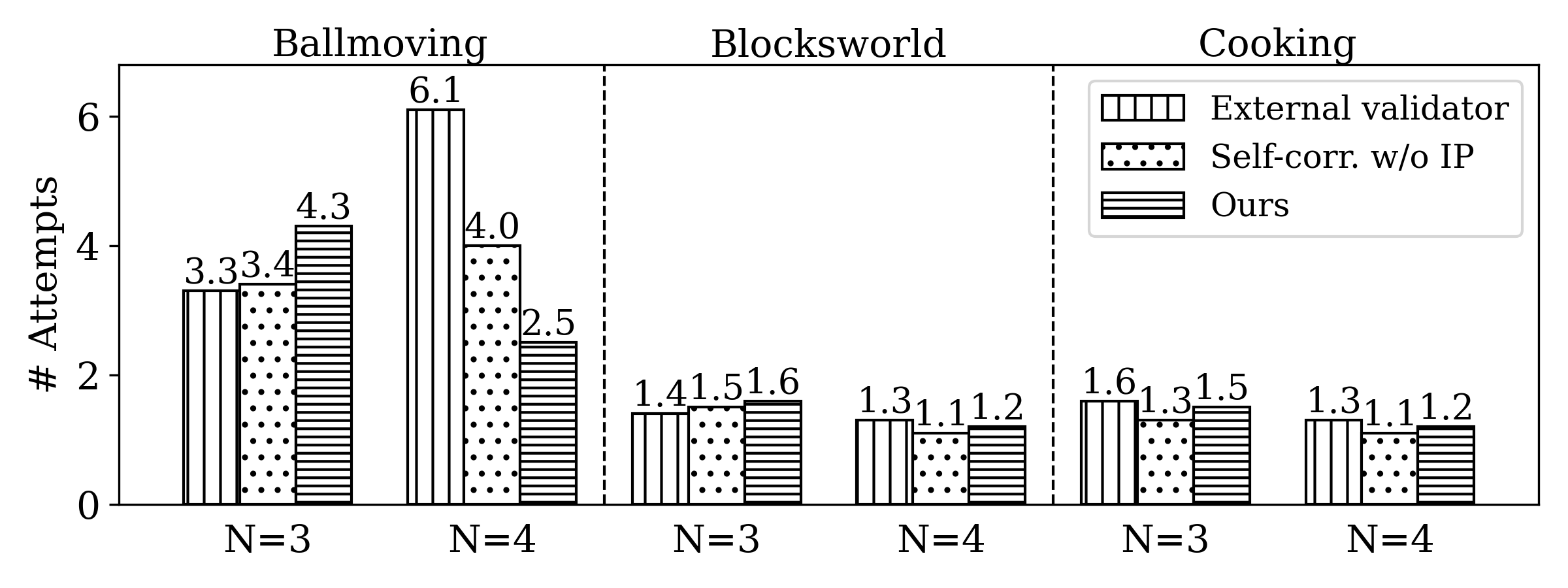}
        \caption{Gemini-1.5-Flash}
        \label{subfig:fig_4_b}
    \end{subfigure}
    \caption{ 
    The number of attempts for correction.}
    \label{fig:fig_4}
\end{figure}

We also measured the number of attempts for the correction, as plotted in Fig.~\ref{fig:fig_4}.
If the result from each correction step is infeasible, a new attempt to refine the original plan is initiated to revise it. 
In most scenarios, the proposed method requires the fewest refinement attempts while achieving the highest performance.
For Ballmoving and Cooking, fewer attempts were required compared to those using external validators.
It suggests that the feedback generated by InversePropmt is more interpretable and accurate than predefined error types and feedback for re-planning.
The self-correction method without the proposed inverse prompting occasionally shows fewer attempts in Cooking and Blocksworld due to its failure to detect errors.
As a result, it fails to make further attempts to correct the previously flawed plan.
%-----------------------------------------------------------------%
%
%
%-----------------------------------------------------------------

Figure~\ref{fig:fig_2} illustrates selected examples of feedback generated by the self-correction process using the proposed inverse prompting and standard prompting methods for Ballmoving.
When employing standard prompting, we often encounter instances of undetected errors and the generation of incorrect feedback.
Specifically, Fig.~\ref{fig:fig_2} shows that standard prompting often produced plausible yet incorrect feedback, such as identifying the incorrect location of {\textit{ball1}} instead of the position of {\textit{robot1}}.
In contrast, the proposed inverse prompting method with its reasoning steps provided detailed feedback on the exact location of the {\textit{robot1}} and offered clear justifications for errors in the previously generated plan.

%-----------------------------------------------------------------%
%
\begin{figure*}[t!]
    \centering
    \begin{subfigure}[t]{1.0\linewidth}
        \centering
        \includegraphics[width=0.98\columnwidth]{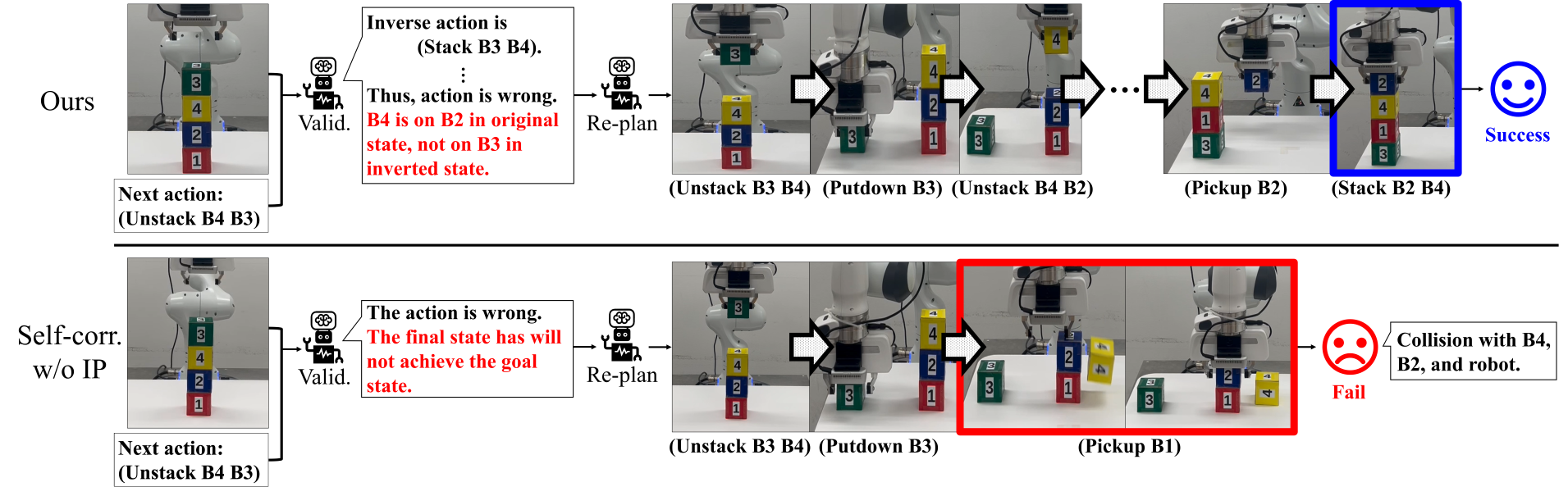}
        \caption{In Blocksworld, the goal is to stack the blocks in the following order: Block3 (\textit{B3}) on the table,  \textit{B1} on \textit{B3}, \textit{B4} on \textit{B1}, and \textit{B2} on \textit{B4}.}
        \label{subfig:fig_5_a}
    \end{subfigure}
    \\\vspace{3mm}
    \begin{subfigure}[t]{1.0\linewidth}
        \centering
        \includegraphics[width=0.98\columnwidth]{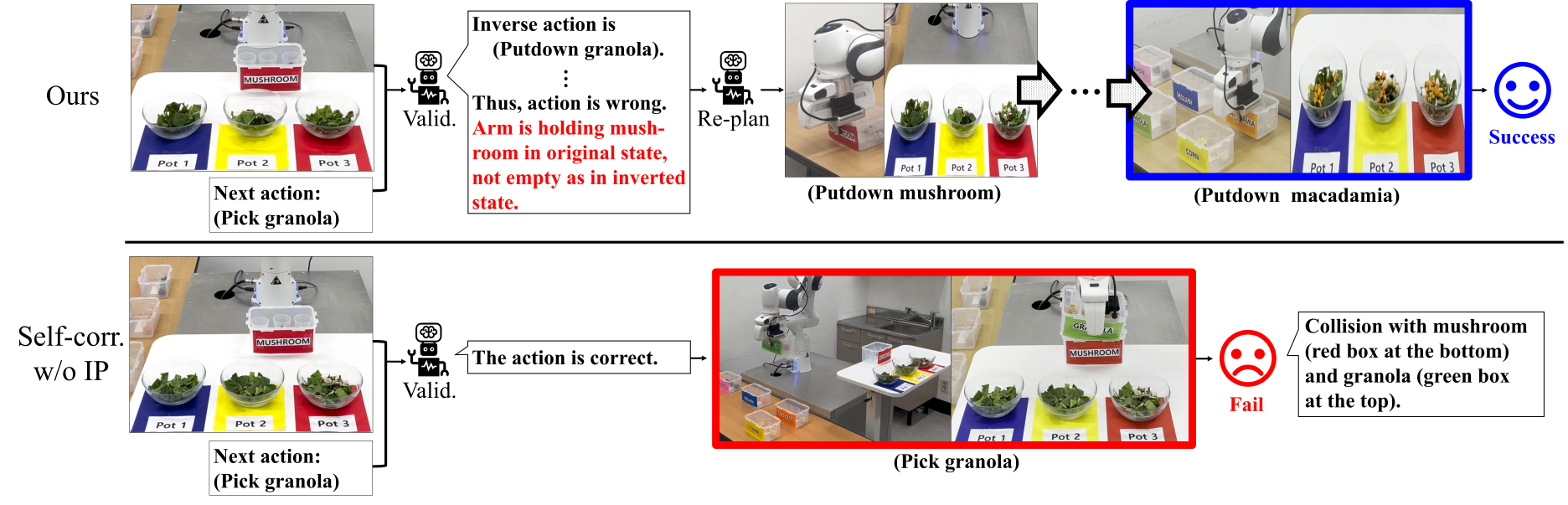}
        \caption{In Cooking, the objective is to distribute the ingredients such that \textit{Pot1} contains \textit{granola} and \textit{carrots}, \textit{Pot2} contains  \textit{granola}, \textit{raisins}, \textit{carrots}, and \textit{macadamias}, and \textit{Pot3} contains \textit{mushroom} and \textit{carrots}.}
        \label{subfig:fig_5_b}
    \end{subfigure}
    \\\vspace{1mm}
    \caption{
    Comparison of execution sequences in real-world scenarios between self-corrective task planning using the proposed method with inverse prompting method (Ours) and without it (Self-corr. w/o IP).
    \protect\includegraphics[height=2.5ex, width =3ex]{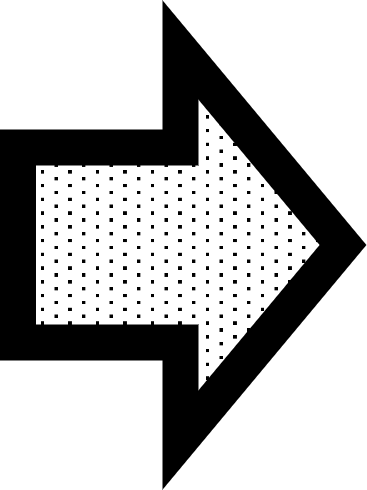} indicates that the given action was determined to be correct in the validation process and was subsequently executed.
    Best viewed in color ($\times$2).}
    \label{fig:fig_5}
\end{figure*}
%
%-----------------------------------------------------------------%

%%%%%%%%%%%%%%%%%%%%%%%%%%%%%%%%%%%%%%%%%%%%%%%%%%%%%%%%%%%%%%%%%%%%%%%%%%%%%%%%
%%%%%%%%%%%%%%%%%%%%%%%%%%%%%%%%%%%%%%%%%%%%%%%%%%%%%%%%%%%%%%%%%%%%%%%%%%%%%%%%
\subsection{Real-world Experiments}
\label{sec:real_world_exp}
%%%%%%%%%%%%%%%%%%%%%%%%%%%%%%%%%%%%%%%%%%%%%%%%%%%%%%%%%%%%%%%%%%%%%%%%%%%%%%%%
\subsubsection{Setup}
We evaluated the proposed method using a 7-DOF Franka Emika Panda robot arm in both the Blocksworld and Cooking scenarios.
In the Blocksworld scenario~\cite{liu2023llm+}, the robot is required to stack four blocks in the specified order.
In the Cooking scenario~\cite{zhou2024isr}, the robot aims to complete various salad recipes by adding ingredients such as carrots, raisins, mushrooms, granola, and macadamias into three pots, initially containing only green vegetables.
We validated the feasibility of the generated action sequence in each environment state through the self-correction process. 
If its results indicate an infeasible action sequence, we replanned it based on the generated feedback.
Otherwise, we executed the robot action for that sequence.
To verify the effectiveness of the proposed inverse prompting approach in the self-correction process, we compared it with existing methods that employ standard prompting~\cite{zhou2024isr,ming2023hicrisp}.

%%%%%%%%%%%%%%%%%%%%%%%%%%%%%%%%%%%%%%%%%%%%%%%%%%%%%%%%%%%%%%%%%%%%%%%%%%%%%%%%
\subsubsection{Results}
Fig.~\ref{fig:fig_5} shows the actual action sequence performed by the robot.
Note that \textit{(action x y)} means that the \textit{action} is performed on the \textit{x} with respect to \textit{y} (sometimes \textit{y} is omitted).
In the Blocksworld, the action initially generated by the planner, \textit{(Unstack B4 B3)}, is infeasible in the current state.
Ours provided a clear explanation with detailed justification for its infeasibility, resulting in the correctly revised action sequence and successful completion of the task.
In contrast, the standard prompting method detected the error but produced unhelpful feedback, leading to the infeasible subsequent plan, \textit{(Pickup B1)}. 
Afterward, the self-correction process failed to detect this error, resulting in task failure due to a collision with B4, B2, and the robot.

In Cooking, InversePrompt provided detailed feedback on the third plan \textit{(Pick granola)} but non-executable, specifying that the arm is holding a mushroom, facilitating successful re-planning and task completion by the robot. 
In contrast, the self-correction method without inverse prompting failed to detect the incorrectly generated plan, \textit{(Pick granola)}, mistakenly validating it as a correct action.
As a result, the robot executed the action, leading to task failure due to a collision with the mushroom in the red box.

%%%%%%%%%%%%%%%%%%%%%%%%%%%%%%%%%%%%%%%%%%%%%%%%%%%%%%%%%%%%%%%%%%%%%%%%%%%%%%%%
%%%%%%%%%%%%%%%%%%%%%%%%%%%%%%%%%%%%%%%%%%%%%%%%%%%%%%%%%%%%%%%%%%%%%%%%%%%%%%%%

\section{CONCLUSIONS}
We have introduced novel self-corrective planning of tasks with an inverse prompting strategy, named InversePrompt, which incorporates multi-step reasoning to provide explicit justifications for feedback. 
It first generates the inverse action of the produced action plan to examine the feasibility and logical flow of the generated action sequence.
Then, it analyzes the inverted state resulting from the inverse action and the original state to generate feedback.
Comprehensive experimental results on benchmark datasets and real-world scenarios demonstrate that the proposed method outperforms existing LLM-based competitors in task completion, self-correction, and feedback generation.
By providing more accurate and informative feedback, it achieves a higher success rate while requiring fewer refinement steps.

%%%%%%%%%%%%%%%%%%%%%%%%%%%%%%%%%%%%%%%%%%%%%%%%%%%%%%%%%%%%%%%%%%%%%%%%%%%%%%%%
%%%%%%%%%%%%%%%%%%%%%%%%%%%%%%%%%%%%%%%%%%%%%%%%%%%%%%%%%%%%%%%%%%%%%%%%%%%%%%%%

\section*{ACKNOWLEDGMENT}
This work was supported in part by Institute of Information \& communications Technology Planning \& Evaluation (IITP) grant funded by the Korea government (MSIT) (2021-0-01341, Artificial Intelligence Graduate School Program(Chung-Ang University)) and in part by a Korea University Grant(K2423001).

%%%%%%%%%%%%%%%%%%%%%%%%%%%%%%%%%%%%%%%%%%%%%%%%%%%%%%%%%%%%%%%%%%%%%%%%%%%%%%%%
%%%%%%%%%%%%%%%%%%%%%%%%%%%%%%%%%%%%%%%%%%%%%%%%%%%%%%%%%%%%%%%%%%%%%%%%%%%%%%%%

\bibliographystyle{IEEEtran}
\bibliography{root}
\end{document}